\documentclass[10pt,twocolumn,letterpaper]{article}

\usepackage[pagenumbers]{cvpr} 

\usepackage{epsfig}
\usepackage{graphicx}
\usepackage{amsmath}
\usepackage{amssymb}
\usepackage[noend]{algpseudocode}
\usepackage{url}
\usepackage{graphicx}
\usepackage{enumitem}
\usepackage{caption}
\usepackage{xcolor}
\usepackage{algorithm}
\usepackage{mathtools}
\usepackage{color, colortbl}
\usepackage{url}
\usepackage{caption}
\usepackage{subcaption}
\usepackage{float}
\usepackage{bbm}

\usepackage{multirow}
\usepackage{pifont}
\usepackage{amsmath,amsfonts,bm}

\def\eqref#1{equation~\ref{#1}}

\def\1{\bm{1}}

\DeclareMathAlphabet{\mathsfit}{\encodingdefault}{\sfdefault}{m}{sl}
\SetMathAlphabet{\mathsfit}{bold}{\encodingdefault}{\sfdefault}{bx}{n}

\def\Vec#1{{\boldsymbol{#1}}}
\def\Mat#1{{\boldsymbol{#1}}}
\definecolor{Lightgray}{gray}{0.92} 
\definecolor{blond}{rgb}{0.98, 0.94, 0.75}
\definecolor{lightblue}{rgb}{0.1, 0.1, 0.75}
\definecolor{aliceblue}{rgb}{0.94, 0.97, 1.0}
\definecolor{cornflowerblue}{rgb}{0.27, 0.51, 0.8}

\definecolor{cvprblue}{rgb}{0.21,0.49,0.74}
\usepackage[pagebackref,breaklinks,colorlinks,allcolors=cvprblue]{hyperref}

\def\paperID{-} 
\def\confName{CVPR}
\def\confYear{2026}

\title{Echoes Over Time: Unlocking Length Generalization in Video-to-Audio Generation Models}

\author{%
Christian Simon$^{\dagger}$ \quad Masato Ishii$^{\clubsuit}$ \quad Wei-Yao Wang$^{\dagger}$ \quad Koichi Saito$^{\clubsuit}$  \\   Akio Hayakawa$^{\clubsuit}$ \quad Dongseok Shim$^{\dagger}$ \quad
  Zhi Zhong$^{\dagger}$ \quad Shuyang Cui$^{\dagger}$ \\ Takashi Shibuya $^{\clubsuit}$  \quad Shusuke Takahashi$^{\dagger}$  \quad Yuki Mitsufuji$^{\dagger, \clubsuit}$  \vspace{0.3cm} \\
  $^{\dagger}$Sony Group Corporation \quad $^{\clubsuit}$Sony AI  \vspace{-0.1cm} \\
  {\fontsize{9}{16}\texttt{\{first\_name.last\_name\}@sony.com}} \vspace{-0.01cm} 
  \\ 
}

\begin{document}
\maketitle

\begin{abstract}
Scaling multimodal alignment between video and audio is challenging, particularly due to limited data and the mismatch between text descriptions and frame-level video information. 
In this work, we tackle the scaling challenge in multimodal-to-audio generation, examining whether models trained on short instances can generalize to longer ones during testing.
To tackle this challenge, we present multimodal hierarchical networks so-called  MMHNet, an enhanced extension of state-of-the-art video-to-audio models. Our approach integrates a hierarchical method and non-causal Mamba to support long-form audio generation. Our proposed method significantly improves long audio generation up to more than 5 minutes. We also prove that training short and testing long is possible in the video-to-audio generation tasks without training on the longer durations. We show in our experiments that our proposed method could achieve remarkable results on long-video to audio benchmarks, beating prior works in video-to-audio tasks. Moreover, we showcase our model capability in generating more than 5 minutes, while prior video-to-audio methods fall short in generating with long durations. Our project page:  \url{https://echoesovertime.github.io}.
\end{abstract}

\section{Introduction}
\label{sec:intro}

\begin{figure}[t]
  \centering
  \includegraphics[width=0.97\linewidth]{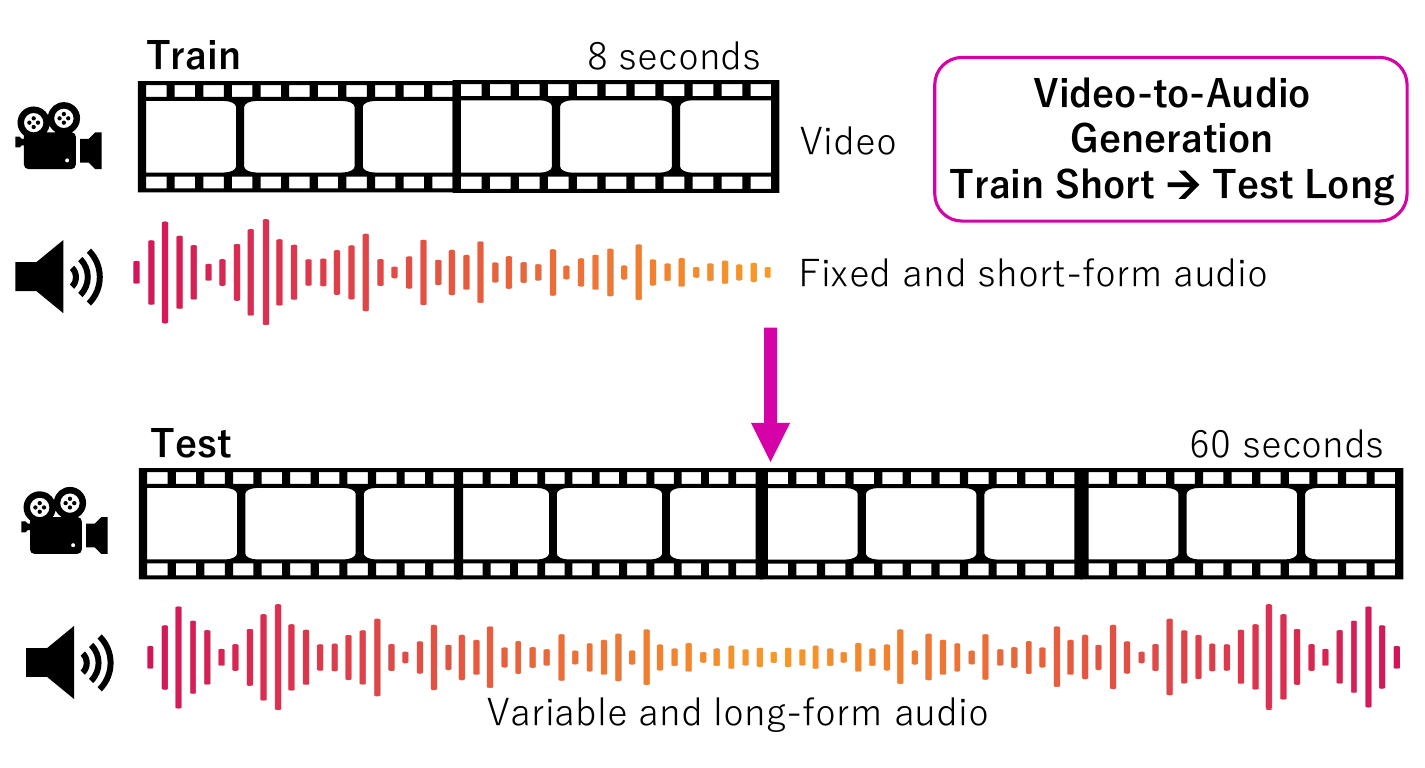}
  \caption{
    {Long-Video to Audio (LV2A) task overview. The challenge is framed as training models on fixed-length segments while requiring them to generalize to variable-length (long-form) audio outputs during inference.}  
  }
  \label{fig:intro}
  \vspace{-0.4cm}
\end{figure}

Video-to-Audio (V2A) is a generative task that aims to produce realistic and contextually aligned audio from silent video inputs. This capability holds substantial promise for enhancing sound design workflows, particularly in domains such as film and gaming~\cite{Li_2024_CVPR,zhang2024audio}. Despite its potential, existing V2A methods~\cite{cheng2025mmaudio,iashin2021tiva,sheffer2023im2wav,luo2023difffoley,zhang2024foleycrafter} are primarily tailored for short-form audio generation, typically spanning 8–10 seconds. Among these, diffusion-based approaches~\cite{cheng2025mmaudio,luo2023difffoley,zhang2024foleycrafter} have shown superior performance over transformer-based autoregressive models~\cite{sheffer2023im2wav}, largely by denoising fixed-length noise segments, which is a strategy well-suited for brief clips.
However, extending these models to long-form video inputs is challenging due to limited training data and the substantial memory requirements for modeling extended audio sequences. 
For instance, on some publicly available long audio-video datasets~\cite{geng2025longvale,geng2023unav100}, the distributions mostly cover only up-to 1 minute video.
When applied to long-video-to-audio (LV2A) tasks, existing models trained on fixed-length segments struggle to accommodate longer sequence generation in testing, thereby constraining their effectiveness in real-world applications.



We are interested in train-short and test-long problems where the longer video duration (up to 5 minutes) could be generated properly using only short clips in our training data as shown in Figure~\ref{fig:intro}. 
Generating short clips for each short duration could be an alternative for LV2A~\cite{zhang2025lvas}. Despite its practicality, this method often results in fragmented audio experiences, marked by disjointed transitions, unaligned sound events, and degraded audio quality stemming from its limited grasp of long-form video context. Please see Sec.~\ref{sec:pilot-study} to see our early observations. 

In particular, we identify that existing V2A models~\cite{cheng2025mmaudio,iashin2021tiva,sheffer2023im2wav,luo2023difffoley,zhang2024foleycrafter,viertola2025vaura} expose structural constraints that reduce the generalizability in terms of various length generation 
and performance. The core base architecture of these models relies on transformer models~\cite{vaswani2017attention}. Thus, these existing models depend on explicit positional encodings that are difficult to tame when dealing with longer sequence generation. Explicit positional encodings often hurt generalization to longer sequences~\cite{kazemnejad2023impact}.
Fortunately, Mamba~\cite{gu2023mamba,dao2024mamba2} is introduced as an alternative to transformer modules, showing strong performance on various tasks and modalities~\cite{hatamizadeh2025mambavision,hu2024zigma,gu2023mamba,dao2024mamba2,shams2024ssamba}. Thus, there is an alternative to avoid using explicit positional encodings, which is deteriorating in generating long outputs. 

To tackle the challenges in LV2A generation, we introduce MMHNet, a novel framework that reconceptualizes the task as one of multimodal alignment across modalities with varying token lengths. Our proposed method could effectively align between modalities and handle long video and audio without further adjustment in the model during inference.
MMHNet combines a multimodal video-to-audio (V2A) model with the HNet architecture~\cite{hwang2025hnet}, enabling audio synthesis conditioned on diverse multimodal inputs while effectively aligning visual and textual modalities.
HNet enhances token processing through a hierarchical structure, moving beyond conventional attention mechanisms. By replacing standard attention blocks with HNet and incorporating dynamic chunking, routing, and smoothing modules, MMHNet achieves effective and coherent audio generation over long durations. Unlike causal models, MMHNet leverages video conditions, which are non-causal, maintaining a global receptive field that supports high-quality audio synthesis for long videos.
Our method operates in a compressed space during early layers, where multimodal alignment occurs to effectively integrate tokens from different sources and reduce redundancy. This approach leverages inherent overlaps in visual and audio data (\eg, similar frames and sound events within the same timeframe)~\cite{xiao_retake_2024,pumbley2009sparserep}.
We introduce multimodal based routing to bridge distinct modalities and apply time-based token routing to reduce temporal complexity and enhance cross-modal alignment.


To evaluate MMHNet’s capabilities, we introduce a long-form V2A evaluation benchmark built upon the UnAV100~\cite{geng2023unav100} and LongVale~\cite{geng2025longvale} datasets. Our experimental results show that MMHNet not only sets a new standard in long-form audio generation but also consistently delivers high-quality outputs across different durations.

The contributions of our work are threefolds:
\begin{itemize}
    \item We introduce the length generalization challenge by training on short, fixed-length audio-visual data and evaluating on long-form video-to-audio (V2A) generation tasks using the UnAV100 and LongVale datasets.
    \item We propose MMHNet, a multimodal hierarchical network that integrates MMAudio and hierarchical networks  for efficient and consistent long-form audio generation.
    \item We conduct extensive experiments across long-form benchmarks, validating MMHNet’s superior performance and ability to scale with video duration.
\end{itemize}

\section{Related work}
\label{sec:related_work}
\noindent \textbf{Video-to-audio generation.}
Video-to-audio synthesis aims to generate sound that is both semantically and temporally aligned with visual content. Existing methods typically fall into two categories: 1) those that inject visual features into pre-trained text-to-audio (TTA) models, and  2) those that train video-to-audio (V2A) models from scratch. Approaches like T2AV~\cite{mo2024t2av} and FoleyCrafter~\cite{zhang2024foleycrafter} enhance visual consistency and alignment by integrating visual and textual embeddings into audio generation pipelines. Meanwhile, models \eg, Diff-Foley~\cite{luo2023difffoley} and Frieren~\cite{wang2024frieren} leverage contrastive pre-training and flow matching to improve multimodal coherence. MMAudio~\cite{cheng2025mmaudio} further advances this field with a hybrid architecture combining multimodal and single-modality diffusion transformer (DiT) blocks~\cite{peebles2023dit}, incorporating synchronization features validated by Synchformer~\cite{iashin2024synchformer} and visual semantic features from CLIP~\cite{radford2021clip}. V-AURA~\cite{viertola2025vaura} is proposed as an autoregressive method to generate audio from given video frames. However, all of these methods are only well-suited for short-form video-to-audio generation, which limits the capability in generating audio beyond the duration covered during training. HunyuanVideo-Foley~\cite{shan2025hunyuanvideofoley} was recently introduced, showcasing strong audio generation capabilities from diverse inputs such as text, SigLIP visual embeddings~\cite{tschannen2025siglip}, and Synchformer~\cite{iashin2024synchformer}. Nevertheless, previous approaches have yet to fully unlock the potential for generating audio beyond the scope of training data.


\begin{figure*}[!t]
  \centering
  \includegraphics[width=1.01\linewidth]{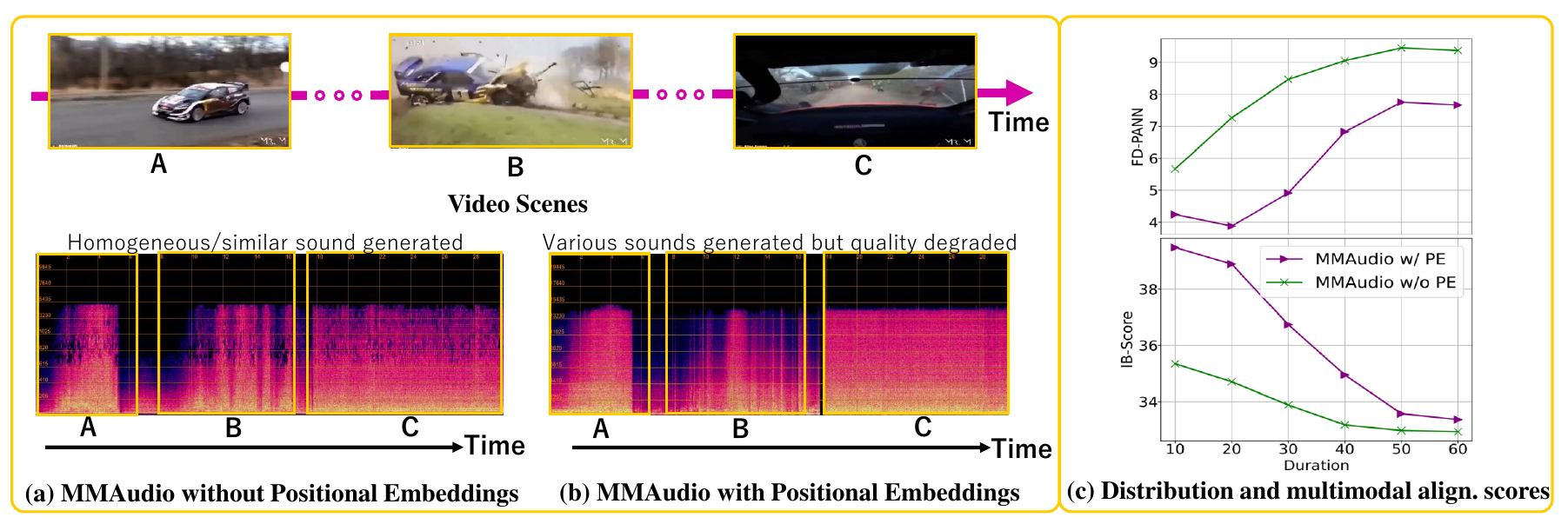}
  \caption{
    {
    We analyze the role of positional embeddings in V2A models such as MMAudio~\cite{cheng2025mmaudio}, built on MMDiT~\cite{flux2024}. Without positional embeddings (a), MMAudio fails to capture temporal structure, producing redundant audio dominated by prominent visual objects (\eg, car crashing). With adjusted positional embeddings (b), alignment improves but sound quality degrades over long sequences (see scene C). (c) On UnAV100~\cite{geng2023unav100}, both configurations show performance drops across durations, with MMAudio without positional embeddings performing worst in distribution matching (FD$_{PANN}\downarrow$) and multimodal alignment (IB-Score$\uparrow$).
  }  
  }
  \label{fig:postemb_analysis}
\end{figure*}

\noindent \textbf{Multimodal models.}
Multimodal conditioning (\eg, video and text) is vital to current generative models~\cite{cheng2025mmaudio,luo2023difffoley,viertola2025vaura,saito2025soundreactor,Simon2025titanguide}, with many V2A systems relying on Transformer architectures for multimodal processing. While Transformers are effective in multimodal tasks~\cite{peebles2023dit}, their dependence on positional embeddings limits generalization to durations beyond training. Position scaling, \eg, NTK or interpolation, is often required to extend their temporal range~\cite{Tancik2020ntk, peng2024yarn, chen2023pirope}. 
In contrast, Mamba~\cite{gu2023mamba,dao2024mamba2} processes sequences without positional embeddings, enabling efficient long-duration generation without  modifications.

\noindent \textbf{Long video-to-audio generation.}
LoVA~\cite{cheng2025lova} represents the current state-of-the-art in long-video-to-audio generation, leveraging DiT-based architectures~\cite{peebles2023dit} to produce coherent and temporally aligned audio tracks from extended video inputs. It significantly outperforms earlier models in generating synchronized and contextually appropriate audio for long-form video content. Despite its strengths, LoVA exhibits limitations when tasked with generating audio beyond the one-minute mark, often resulting in noticeable degradation in audio quality and coherence. Autoregressive models~\cite{viertola2025vaura,sheffer2023im2wav} offer an alternative approach, showing promise in long-form generation due to their step-by-step prediction capabilities. However, they are prone to error accumulation over time, which can lead to drift and loss of fidelity in extended sequences. Another promising direction involves agent-based methods~\cite{zhang2025lvas}, which divide long videos into shorter, manageable segments and generate audio for each clip independently. While this segmentation strategy can improve scalability and maintain quality, it introduces additional complexity by requiring accurate text descriptions for each segment and precise control over clip transitions to ensure seamless audio continuity.

\section{Pilot Study}
\label{sec:pilot-study}
\noindent \textbf{Why do Transformer-based V2A models fail to
generalize to long sequences?} We observe that certain aspects of the current Transformer architecture in V2A, specifically positional embeddings~\cite{chen2023pirope} and attention logit exploding~\cite{han2024lminfinite}, pose challenges to length generalization unless substantial modifications are made in the inference mode.

\noindent \textbf{Problems with positional embeddings.} 
Positional embeddings like RoPE~\cite{su2024roformer} are essential for Transformer-based models, as they provide positional awareness for tokens. Without them, the model loses this capability. Training without positional embeddings is only viable when training and testing use identical sequence lengths. To explore this, we conduct a pilot study analyzing a pretrained Transformer-based model (\eg, MMAudio~\cite{cheng2025mmaudio}) trained on 8-second audio-visual data and tested on longer sequences (\eg, 40 seconds), As illustrated in Figure~\ref{fig:postemb_analysis}, pretrained video-to-audio models without positional embeddings would perform poorly. Also, the generated sound becomes homogeneous for the model without positional embeddings because attention modules are orderless and the semantic meaning becomes less on point relative to the positions as shown in Figure~\ref{fig:postemb_analysis} (a). Figure~\ref{fig:postemb_analysis} (c) shows that increasing durations degrade the Transformer based V2A model significantly, 3-4 points drop for distribution matching (FD$_{PANNs}$) and multimodal alignment (IB) scores. Designing a network without positional embeddings is preferable in this case to avoid unnecessary adjustments when generating longer sequences during testing. 



\section{Proposed Method}
Let $\mathcal{D}$ be a dataset where each sample $(\Vec{x}, \Vec{c}) \in \mathcal{D}$ comprises an audio $\Vec{x}$ and associated conditions $\Vec{c}$ (\eg, video frames and a text caption). The objective is to train a model on $\mathcal{D}$ to learn a conditional distribution $p_{\text{model}}(\Vec{x} \mid \Vec{c})$ that closely approximates the true data distribution $p_{\text{data}}(\Vec{x} \mid \Vec{c})$ via flow matching~\cite{liu2022flow,lipman2022flow}. Our focus lies particularly on scenarios where $\Vec{x}$ represents a long-form audio, significantly exceeding the lengths typically handled by existing methods, which often operate on short clips of approximately 10 seconds during both training and inference.

To effectively model long-form audio distributions, we design the core architecture using Mamba-2 variants~\cite{dao2024mamba2,shi2024vssd}, which enable token processing without relying on positional embeddings. This choice is motivated by our observation in Sec.~\ref{sec:pilot-study} that positional embeddings tend to degrade performance in long audio generation scenarios. Additionally, to enhance cross-modal alignment, we incorporate routing strategies that reduce token redundancy by compressing repetitive information, thereby improving efficiency and coherence across modalities.


\subsection{Preliminaries}
\noindent \textbf{Flow matching.}
We employ the conditional flow matching objective~\cite{lipman2022flow,liu2022flow} for generative modeling. For detailed methodology, we refer readers to~\cite{lipman2022flow}. Briefly, during inference, a sample is generated by first drawing noise $\Vec{x}_0$ from a standard normal distribution. An ODE solver is then used to numerically integrate from time $t = 0$ to $t = 1$, guided by a learned, time-dependent conditional velocity vector field: $
v_{{\theta}}(t, \Vec{c}, \Vec{x}): [0, 1] \times \mathbb{R}^C \times \mathbb{R}^d \rightarrow \mathbb{R}^d,
$
where   $\Vec{c}$ denotes the conditioning input (\eg., video and text), and $\Vec{x}$ is a point in the vector field. The velocity field is parameterized by a deep neural network with parameters ${\theta}$.
At training time, we learn the parameters $\theta$ of the deep neural network by minimizing the following objective:
\begin{equation}
\mathbb{E}_{t, (\Vec{x}_0, \Vec{x}_1, \Vec{c}) \sim q(\Vec{x}_0) q(\Vec{x}_1, \Vec{c})} \left[ \left\| v_{\theta}(t, \Vec{c},\Vec{x}_t) - u(\Vec{x}_t \mid \Vec{x}_0, \Vec{x}_1) \right\|^2 \right],
\end{equation}
where $t \sim \mathcal{U}[0, 1]$ is sampled uniformly from the interval $[0, 1]$, and $q(\Vec{x}_0) q(\Vec{x}_1, \Vec{c})$ denotes the joint distribution over the prior and training data.
The interpolated point $\Vec{x}_t$ is defined as:
\begin{equation}
\Vec{x}_t = t \Vec{x}_1 + (1 - t) \Vec{x}_0,
\end{equation}
and the corresponding flow velocity at $\Vec{x}_t$ is given by:
\begin{equation}
    u(\Vec{x}_t \mid \Vec{x}_0, \Vec{x}_1) = \Vec{x}_1 - \Vec{x}_0.
\end{equation}
Our model is designed to predict the flow over $T$ steps during training. To ensure efficiency and practicality in sample generation, we perform flow matching within the latent space.




\subsection{Base Architecture}
\noindent \textbf{Multimodal (MM) flow-matching model.}
To support multimodal generation, we adopt the MMAudio~\cite{cheng2025mmaudio} model structure following the MM-DiT block architecture from SD3~\cite{esser2024sd3} and FLUX~\cite{flux2024} with multiple streams of modilities and single-modality blocks.
This design choice enables us to construct deeper networks without increasing the overall parameter cost, compared to architectures that process all modalities at every layer.
This multimodal architecture allows the model to dynamically attend to different modalities based on the input context, thereby enabling efficient joint training on both audio-visual and audio-text datasets. 
\paragraph{Multimodal conditioning inputs.} 
To incorporate global context into the network, we adopt global conditioning via adaptive layer normalization (adaLN)~\cite{perez2018film}, where global features are injected through learned scale and bias parameters. Specifically, we compute a global conditioning vector $\Vec{c}_g \in \mathbb{R}^{1 \times D}$, which is shared across all Transformer blocks with average-pooled visual and text features.
To further enhance audio-visual synchrony, we also employ token-level conditioning, allowing the model to adapt more precisely to local variations across modalities. In our implementation, we make use of semantic video representation from CLIP~\cite{radford2021clip}, motion-audio synchronized representations from Synchformer~\cite{iashin2024synchformer}, and the text representations from CLIP~\cite{radford2021clip}.

\begin{figure*}[t]
  \centering
  \includegraphics[width=.88\linewidth]{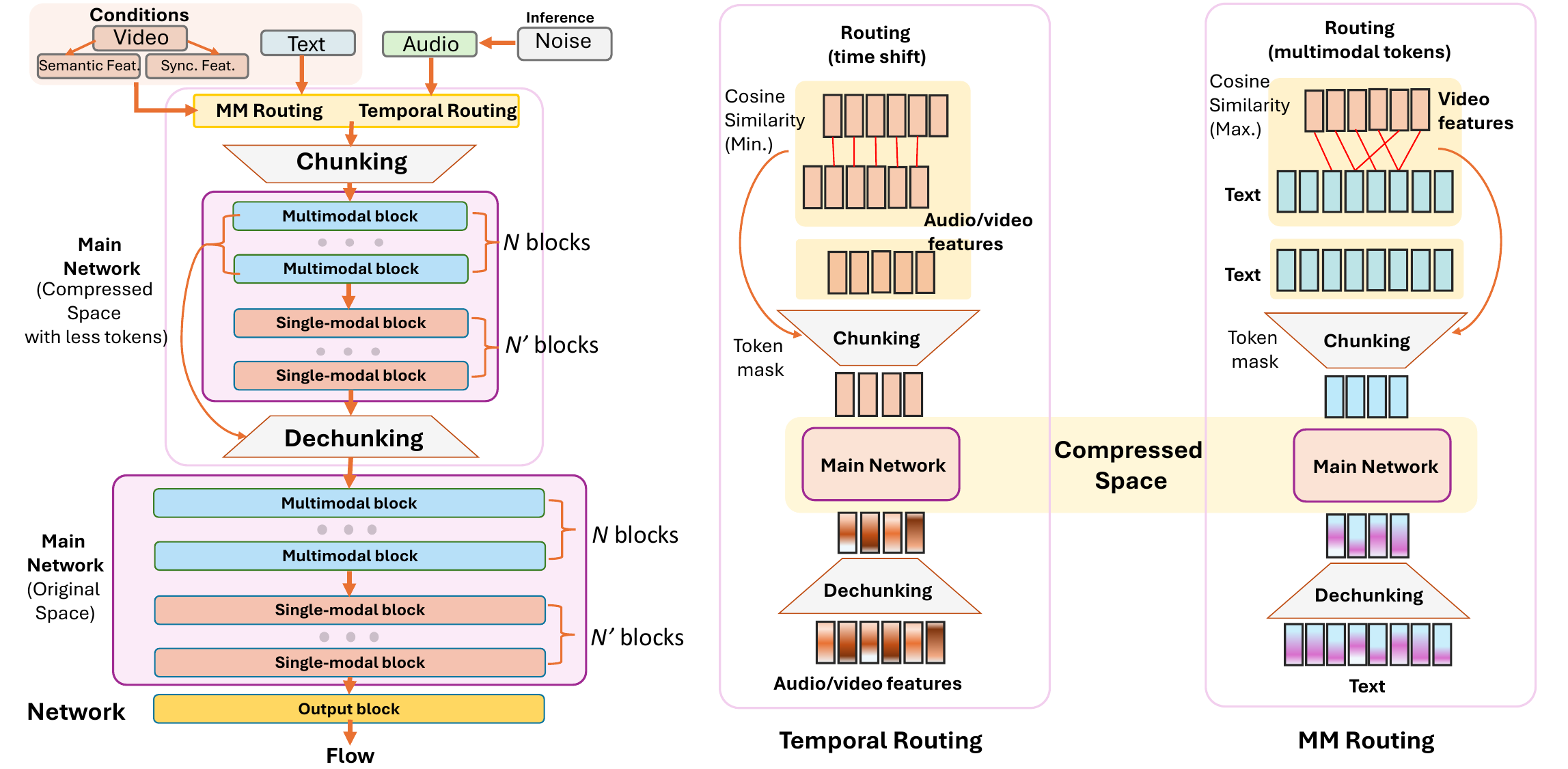}
  \caption{
    {Overview of our proposed framework. Left: A comprehensive end-to-end flow-matching model that operates across both multimodal and single-modal blocks, handling inputs in both compressed and original spaces. Middle: A temporal routing mechanism designed to efficiently process tokens in a time-aware manner. Right: A multimodal routing strategy that leverages strong correlations between the two modalities for enhanced integration.}  
  }
  \label{fig:mmhnet_architecture}
\end{figure*}

\subsection{Core Network}
\noindent \textbf{No positional embeddings by replacing attention modules in single modality blocks.} 
Traditional attention mechanisms in Transformers~\cite{vaswani2017attention} face significant challenges when applied to long-form audio generation. These modules rely heavily on positional embeddings to compute attention scores between queries and keys. However, such embeddings are typically fixed during training and do not generalize well when the number of tokens changes at inference time, leading to degraded performance on longer sequences. This limitation necessitates a more adaptable modeling approach that can handle variable-length inputs without relying on rigid positional encoding schemes.
To address this, we adopt the Mamba-2 architecture~\cite{dao2024mamba2}, which inherently supports sequence modeling without explicit positional embeddings. Mamba-2 leverages a state-space model formulation, where adaptive tokens provide contextual information to a transition matrix, enabling the model to capture temporal dependencies across token sequences more flexibly. This design allows for robust generalization to longer sequences not seen during training, without requiring architectural modifications or extrapolation techniques, as is often necessary with Transformer-based models using rotary positional embeddings (RoPE)~\cite{su2021roformer}.

Here, we briefly introduce Mamba-2's parameterization~\cite{dao2024mamba2} as our basic model in this work. Let ${x}_i, {y}_i \in \mathbb{R}$ be the input and output of the state space model (SSM), respectively. The model is parameterized by: $\Vec{A} \in \mathbb{R}_{<0}, \Mat{B} \in \mathbb{R}^{n}$, and $\Mat{C} \in \mathbb{R}^{n}$. Then, the discretization is written as $\alpha_\ell=e^{\Delta_\ell A_\ell} \in (0,1)$ and $\gamma_\ell = \Delta_\ell$.
We formally define the SSM in Mamba: 
\begin{align}
\Vec{h}_{\ell} = \alpha_\ell \Vec{h}_{\ell-1} + \gamma_\ell \Vec{B}_\ell x_\ell, \;\;\;\;\;{y}_\ell = \Vec{C}^\top \Vec{h}_\ell
\end{align}
with the matrix form as follows: 
\begin{align}
\Vec{Y} = \Big( \Vec{M} \odot \Vec{C} \Vec{B}^\top \Big)\Vec{X}, 
\end{align}
where $\Vec{M} \in \mathbb{R}^{L \times L}$ is the structured mask matrix consisting of $\alpha_\ell$, $\Mat{B}, \Mat{C} \in \mathbb{R}^{L \times N}, \Mat{X} \in \mathbb{R}^{L \times D}$ are the SSM parameters and inputs, respectively.


\noindent \textbf{Non-Causal Mamba-2 modules.} 
We adopt Non-Causal Mamba-2~\cite{shi2024vssd} for two key reasons: 1) video conditions are available offline, eliminating the need for sequential token processing, and 2) multimodal fusion across multiple modalities is difficult without a predefined order. The original Mamba-2~\cite{dao2024mamba2}, being causal, restricts information flow to one direction, requiring multiple passes to integrate modalities and complicating temporal alignment.
Non-Causal Mamba-2 addresses these limitations by enabling omnidirectional information flow, allowing global hidden states to combine all modalities simultaneously without constrained by scanning orders. 
Non-Causal Mamba-2 also mitigates modulation decay, a common issue in causal models where conditioning signals weaken over time providing a more robust and flexible foundation for multimodal fusion in long-form generation tasks~\cite{ye2025longmamba}.
The key distinction between causal and non-causal Mamba-2 lies in the formulation of the structured mask matrix $\Vec{M}$.
In causal Mamba, $\Vec{m}$ incorporates the product of transformation matrices across the sequence, expressed as $\Vec{A}_{\ell:i} = \prod_i^\ell \Vec{A}_i$. This sequential multiplication leads to decaying~\cite{ye2025longmamba} over long sequences.
In contrast, non-causal Mamba defines $\Vec{m}$ using the inverse of each transformation matrix: $\Vec{m} = \frac{1}{\Vec{A}_i}$. By avoiding cumulative products over time, non-causal Mamba does not experience the same decay phenomenon, making it more stable for long-range dependencies.

\subsection{Hierarchical Framework}
Long video and audio recordings often include a significant amount of redundant information, which can lead to inefficiencies when processing with a large number of tokens, especially in tasks involving multimodal alignment. To address this challenge, we propose a hierarchical framework designed to selectively route only the most important tokens to the main processing network, thereby reducing computational load while preserving critical information.
For example, in the case of audio streams, we implement temporal routing that focuses on identifying the specific timeframes where sound events actually occur. This approach effectively filters out redundant audio data, which is especially useful in scenarios where audio and video streams need to be synchronized, as these streams often contain overlapping or repetitive content.
Furthermore, for multimodal processing, we introduce a multimodal (MM) routing mechanism that selects key tokens based on high similarity between the two modalities \eg, audio and visual data. This selective routing ensures that only the most relevant and informative tokens are passed forward, facilitating more efficient and accurate multimodal alignment.

\noindent \textbf{Routing mechanism.}
We define a routing mechanism based on similarity between two sets of tokens $\Vec{Q} \in \mathbb{R}^{L^\prime\times D}$ and $\Vec{K} \in \mathbb{R}^{L\times D}$, with the similarity function for each instance $\Vec{q}$ and $\Vec{k}$ defined as:
\begin{align}
    \texttt{sim}(\Vec{q}, \Vec{k}) &=   \frac{\Vec{q} ^\top \Vec{k}}{\|\Vec{q}\| \|\Vec{k}\|}  ,  
\end{align}
where this similarity function is used in temporal routing and MM routing layers.

\vspace{0.1cm}
\noindent \textbf{Temporal routing layers.}
In temporal data \eg, audio and video events, the boundaries occur when there are contextual shifts between sound events. Based on this observation, we opt to mask tokens that have high similarities and keep the tokens that contain distinct temporal information. Let $\Vec{q}_\ell = \Mat{W}_q \Vec{x}_\ell$ and $\Vec{k}_\ell = \Mat{W}_k \Vec{x}_\ell$, we use cosine similarity in computing token selection: 
\begin{align}
p_\ell = \frac{1}{2} \Big( 1 - \texttt{sim}(\Vec{q}_\ell, \Vec{k}_{\ell-1}) \Big).
\end{align}

\vspace{0.1cm}
\noindent \textbf{MM routing layers.}
Multimodal alignment between one and another modality (\ie, $M$ and $M^\prime$) might experience deteriorating behavior due to a large number of tokens to be processed. Selected important tokens for feed forwarding to main networks are tokens with high similarity to the referenced modality. For instance, synchronized audio-visual (\ie Synchformer~\cite{iashin2024synchformer}) features could be used to align with text condition. Let $\Vec{q}_{M_\ell}  = \Mat{W}_q \Vec{x}_{M_\ell}$ and $\Vec{k}_{M^\prime_\ell} = \Mat{W}_k \Vec{x}_{M^\prime_{\ell^\prime}}$, we compute MM routing as follows:
\begin{align}
p_\ell = \frac{1}{2} \Big( 1  + \texttt{sim}(\Vec{q}_{M_\ell}, \Vec{k}_{M^\prime_{\ell^\prime}})\Big).
\end{align}
We only process tokens with $b_\ell  = \mathbbm{1}_{\{\texttt{sim}(\Vec{q}_\ell, \Vec{k}_{\ell^\prime}) \geq 0.5\}}$. As conditions are from the pretrained models (\eg, Synchformer~\cite{iashin2024synchformer}, visual CLIP~\cite{radford2021clip}, and text CLIP~\cite{radford2021clip} ), we expect a higher probability for token matching scores (\ie, $>$ 0.5).



\vspace{0.1cm}
\noindent \textbf{Chunking with downsampling.} The downsampler compresses encoder outputs \(\Vec{x}_s\) into a reduced set of vectors \(\Vec{x}_{s+1}\) using boundary indicators \(\{b_{s,\ell}\}_{\ell=1}^{L_s}\). Among potential compression strategies, we adopt direct selection of boundary-marked vectors because of simplicity and effectiveness as suggested in HNet~\cite{hwang2025hnet}.

\noindent \textbf{Dechunking with upsampling.}
After the tokens are processed through the main network, we could obtain output tokens $\tilde{\Vec{x}}$. 
The upsampler is specifically designed to decompress tokens of smaller size back to their original dimensions, enabling more details
processing in the later stages. We define the dechunking with upsampling as follows:
\begin{align}
\Vec{a}_\ell &= p_\ell^{b_\ell} (1 - p_\ell)^{1 - b_\ell} = 
\begin{cases}
p_\ell, & \text{if } b_\ell = 1, \\
1 - p_\ell, & \text{otherwise}.
\end{cases}\\
\end{align}
Then, we make use of Straight-Through Estimator (STE)~\cite{bengio2013estimating}, allowing gradient flow and stop for selected and unselected tokens  $\textrm{STE}(\Vec{a}_\ell) = \Vec{a}_\ell + \textrm{stopgrad}(1 - \Vec{a}_\ell)$, and the output tokens at position $\ell$ could be expressed
$\tilde{\Vec{x}}_\ell = \tilde{\Vec{x}}_{\sum_{k=1}^{\ell} b_k}$. Next, the upsampling function can be defined as:
$\textrm{Upsampler}({\tilde{\Vec{x}}}, \Vec{a})_\ell = \textrm{STE}(\Vec{a}_\ell) \cdot {\tilde{\Vec{x}}}_\ell$.


\section{Experiments}

\begin{table*}[t!]
 \centering
    \resizebox{0.97\textwidth}{!}{
    \Large\addtolength{\tabcolsep}{-2pt}
\begin{tabular}{lccccccccccccc}
\toprule
\rowcolor{Lightgray} \multicolumn{12}{c}{\textbf{UnAV100}} \\
\hline
\multirow{2}{*}{\textbf{Method}} & \multirow{2}{*}{\textbf{Size}} & \multicolumn{4}{c}{Distribution Matching} & & \multicolumn{2}{c}{Audio Quality} & & MM Align. & Temporal Align. \\
\cline{3-6} \cline{8-9} \cline{11-12}
  &  & \textbf{FD$_\textrm{VGG}$ $\downarrow$} & \textbf{FD$_\textrm{PANNs}$ $\downarrow$} & \textbf{FD$_\textrm{PASST}$ $\downarrow$} & \textbf{KL$_\textrm{PASST}$ $\downarrow$} & &\textbf{ISC$_\textrm{PANNs}$ $\uparrow$} & \textbf{ISC$_\textrm{PASST}$ $\uparrow$} & & \textbf{IB-Score $\uparrow$} & \textbf{DeSync $\downarrow$} \\
\toprule
MMAudio-S~\cite{cheng2025mmaudio} & 157M & 4.62 & 10.69 & 349.11 & 1.85 & & 5.92 & 6.02 &&  28.63 & 0.906\\ 
MMAudio-L~\cite{cheng2025mmaudio} & 1.03B & 3.86 & 9.01 & 296.79 & 1.79 & &6.18 & 6.41 &&  30.71 & 0.593 \\
MMAudio-L + NTK~\cite{cheng2025mmaudio} &1.03B  & 4.08 & 8.43 & 301.45 & 1.76 & &6.05 & 6.17 &&  30.24 & 0.599 \\
LoVA~\cite{cheng2025lova} & 1.06B & 3.36 &7.50  & 223.29 & 1.56 & &6.17 & \textbf{8.46} &&  24.62 & 1.232 \\
V-AURA~\cite{viertola2025vaura} & 695M & 4.57 & 6.16 & 321.64 & 1.69 & &7.35 & 5.22 &&  29.36 & 1.191 \\
HunyuanVideo-Foley-XXL~\cite{shan2025hunyuanvideofoley} & 5.13B   & 4.89   &10.28     &284.36    &1.80 & &5.60  &6.01    &&  32.90   & 0.757\\
 \rowcolor{aliceblue} MMHNet - S (ours) & 157M & {3.35} & {5.87} & 217.00 & \textbf{1.29} & & 7.62 & 8.21 &&  36.82 & 0.439 \\
 \rowcolor{aliceblue} MMHNet - L (ours) & 1.09B & \textbf{1.80} & \textbf{5.29} & \textbf{209.06} & 1.49 & &\textbf{8.10} & {7.35} &&  \textbf{36.27} & \textbf{0.410} \\
\toprule
\rowcolor{Lightgray} \multicolumn{12}{c}{\textbf{LongVale}} \\
\hline
\multirow{2}{*}{\textbf{Method}}& \multirow{2}{*}{\textbf{Size}} & \multicolumn{4}{c}{Distribution Matching} & & \multicolumn{2}{c}{Audio Quality} & & MM Align. & Temporal Align. \\
\cline{3-6} \cline{8-9} \cline{11-12}
  &  & \textbf{FD$_\textrm{VGG}$ $\downarrow$} & \textbf{FD$_\textrm{PANNs}$ $\downarrow$} & \textbf{FD$_\textrm{PASST}$ $\downarrow$} & \textbf{KL$_\textrm{PASST}$ $\downarrow$} & &\textbf{ISC$_\textrm{PANNs}$ $\uparrow$} & \textbf{ISC$_\textrm{PASST}$ $\uparrow$} & & \textbf{IB-Score $\uparrow$} & \textbf{DeSync $\downarrow$} \\
\toprule
MMAudio-S~\cite{cheng2025mmaudio} & 157M & 8.86 & 22.23 & 550.57 & 2.34 && 3.27 & 3.42 && 21.52  & 0.972  \\
MMAudio-L~\cite{cheng2025mmaudio} & 1.03B & 7.20 & 16.12 & 531.55 & 2.05 && 2.72 & 2.59 && 21.60  & 0.678  \\
MMAudio-L + NTK~\cite{cheng2025mmaudio} &1.03B  & 6.41 &  13.76 & 481.45 & 2.05   && 2.84  & 2.76  && 23.43  & 0.666 \\
LoVA~\cite{cheng2025lova} & 1.06B & 7.62  &21.81 & 527.58 & 2.36  && 2.46 & \textbf{3.57}  && 17.04 & 1.233   \\
V-AURA~\cite{viertola2025vaura} & 695M &6.46  &14.87   &498.74   &1.87 &&3.22 & 2.67  && 19.67 &1.282 \\
HunyuanVideo-Foley-XXL~\cite{shan2025hunyuanvideofoley} & 5.13B   &14.56   & 28.00     &750.96    & 2.58 &&2.28  &2.40&    &18.75    &1.082 \\
 \rowcolor{aliceblue} MMHNet - S (ours) & 157M & 3.35 & 10.10 & \textbf{323.39}
 & {1.75} & & 3.68 & {3.50} &&\textbf{30.62}  & \textbf{0.438}   \\
 \rowcolor{aliceblue} MMHNet - L (ours) & 1.09B & \textbf{3.23} & \textbf{10.03} & 331.75 & \textbf{1.64} && \textbf{4.25} & 3.20 && 30.00  & 0.465   \\
\hline
\end{tabular}
}
\vspace{-0.1cm}
\caption{Comparison of methods across various evaluation metrics on UnAV100~\cite{geng2023unav100} and LongVale~\cite{geng2025longvale}. 
}
\vspace{-0.2cm}
\label{tab:result_unav100_longvale}
\end{table*}

\noindent \textbf{Settings.}
In our evaluation of long-form audio generation capabilities, we adopt a methodology where the model is initially trained using audio clips of a fixed, relatively short duration, specifically, segments lasting 8 seconds. After this training phase, we rigorously test the model’s ability to generalize by presenting it with much longer audio sequences, each exceeding the original 8-second length. We set multimodal blocks $N=5$ and single modal blocks $N^\prime=4$ for the small version (S), and we use $N=10$ and $N^\prime=7$  for the large version (L). Please see our supplementary materials for the detail architecture and setup, 

\vspace{0.1cm}
\noindent \textbf{Datasets.} We train on VGGSound~\cite{chen2020vggsound} on 8 second audio-video data and several text-to-audio datasets. This datasets have been widely used by our comparing methods. In our experiments, we evaluate on UnAV100~\cite{geng2023unav100} and LongVale~\cite{geng2025longvale} for comparing with the state-of-the-arts on LV2A generation. The test set of UnAV100 consists of $\sim$2K videos with durations of 10-60 seconds, and LongVale has around 1K test videos ranging from 10 to 500 seconds. For completion, we also evaluate on the VGGSound dataset.

\vspace{0.1cm}
\noindent \textbf{Baselines.} 
To demonstrate the effectiveness of our approach in LV2A scenarios, we compare it against LoVA~\cite{cheng2025lova}, a recent method specifically designed for LV2A tasks. Additionally, we evaluate our method against the original MMAudio~\cite{cheng2025mmaudio}, incorporating a frequency scaling of positional embeddings based on the given durations and Neural Tangent Kernel (NTK)~\cite{Tancik2020ntk}. From a conceptual standpoint, autoregressive models are inherently capable of generating longer video sequences by leveraging context window shifts. To assess this capability, we include a comparison with V-AURA~\cite{viertola2025vaura}. We also compare with recent V2A model so-called HunyuanVideo-Foley~\cite{shan2025hunyuanvideofoley}.  

\vspace{0.1cm}
\noindent \textbf{Evaluation on audio-video forms.}  
We evaluate our model across four key dimensions: \textit{distribution matching}, \textit{audio quality}, \textit{semantic consistency}, and \textit{temporal synchronization}. Previous metrics are only feasible for a relatively short audio duration. In our experiments, we conduct an evaluation based on multiple chunks of the audio to match the duration on which the pretrained classifier models are trained. This is to reduce errors where the classifier models cannot directly be applied to long audio-video forms. 

\vspace{0.1cm}
\noindent \textbf{Distribution matching.}  
To measure how closely the generated audio matches the statistical properties of real audio, we compute the Fréchet Distance (FD) and Kullback–Leibler (KL) divergence using established audio embedding models. Specifically, we report FD scores using VGGish~\cite{gemmeke2017vggish} ($\text{FD}_{\text{VGG}}$), PaSST~\cite{koutini2021passt} ($\text{FD}_{\text{PaSST}}$), PANNs~\cite{kong2020panns} ($\text{FD}_{\text{PANNs}}$). PaSST operates at 32 kHz and produces global features, while PANNs and VGGish operate at 16 kHz, with VGGish processing non-overlapping 0.96-second segments. KL divergence is computed using PANNs ($\text{KL}_{\text{PANNs}}$) and PaSST ($\text{KL}_{\text{PaSST}}$) as classifiers.


\vspace{0.1cm}
\noindent \textbf{Audio quality, semantic consistency, and temporal synchronization.}  
We assess the standalone quality of generated audio using the Inception Score (IS), with PANNs~\cite{kong2020panns} serving as the classifier.
To evaluate how well the generated audio semantically aligns with the input video, we use ImageBind~\cite{girdhar2023imagebind} to extract cross-modal embeddings. The cosine similarity between visual and audio features is averaged to yield the \textit{IB-score}.
We evaluate audio-visual alignment using the \textit{DeSync} score, which estimates the temporal offset (in seconds) between audio and video streams. This is computed using Synchformer~\cite{iashin2024synchformer}, a model trained to predict synchronization errors. We assess alignment over the full 4.8-second context window following~\cite{cheng2025mmaudio}.

\begin{table}[t!]
 \centering
    \resizebox{0.497\textwidth}{!}{
    \Large\addtolength{\tabcolsep}{-2pt}
\begin{tabular}{lccccc}
\toprule
\rowcolor{Lightgray} \multicolumn{6}{c}{\textbf{VGGSound}}\\
\textbf{Method}  & \textbf{FD$_\textrm{VGG}$ $\downarrow$}    & \textbf{ISC$_\textrm{PANNs}$ $\uparrow$} & \textbf{ISC$_\textrm{PASST}$ $\uparrow$} & \textbf{IB-Score $\uparrow$} & \textbf{DeSync $\downarrow$} \\
\hline
MMAudio-S~\cite{cheng2025mmaudio}  & 1.66      & 18.02  &  12.46 & 32.27 &0.444 \\
MMAudio-L~\cite{cheng2025mmaudio}  & \textbf{0.97}     & 17.40  & 13.33 & \textbf{33.22} &\textbf{0.442} \\
LoVA~\cite{cheng2025lova}  & 1.70     &9.73  &9.91 &- &- \\
V-AURA~\cite{viertola2025vaura}  & 2.88       &10.08  &- &27.64 &0.654 \\
 \rowcolor{aliceblue}MMHNet-S (ours)  & 1.54     & 16.73  & 12.52 & 32.11 &0.460 \\
 \rowcolor{aliceblue}MMHNet-L (ours)  & 2.09 &\textbf{20.52}  &\textbf{15.34} &32.11 &0.460 \\
\bottomrule
\end{tabular}
}
\vspace{-0.25cm}
\caption{Comparison of methods under a fixed audio length $\sim$10 seconds on VGGSound. Baseline results are based on the reports in~\cite{cheng2025mmaudio,cheng2025lova}.}
\vspace{-0.4cm}
\label{tab:vggsound}
\end{table}

\begin{figure}[t!]
    \centering
    \includegraphics[width=0.444\textwidth]{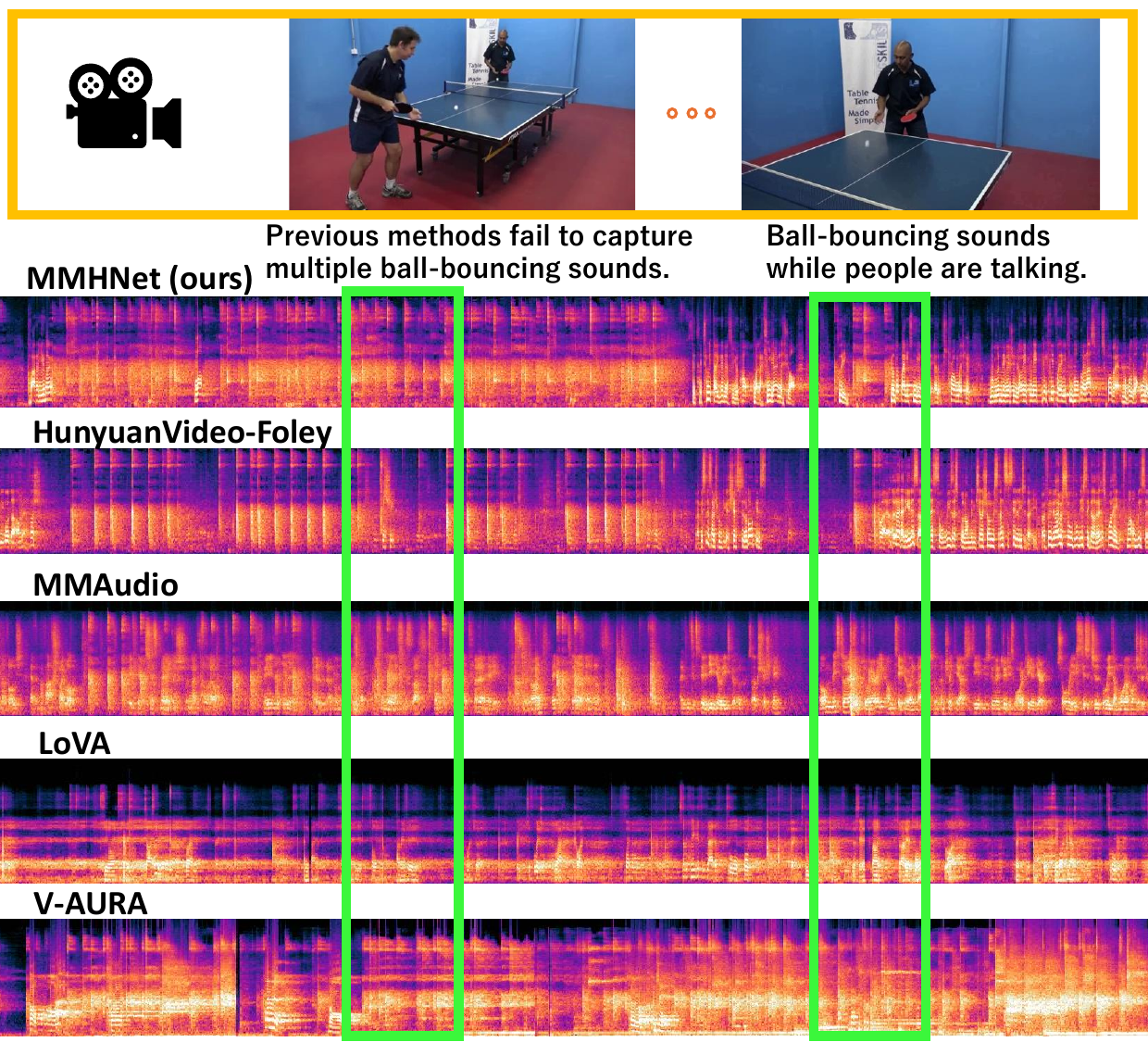}
\vspace{-0.18cm}
\caption{Visualization of audio spectogram from MMHNet and competing methods on UnAV100.}
\vspace{-0.48cm}
     \label{fig:audio_specto}    
\end{figure}

\subsection{Comparison with the state-of-the-arts}
As shown in Table~\ref{tab:result_unav100_longvale}, our proposed model significantly outperforms existing state-of-the-art methods across a broad spectrum of evaluation metrics. In particular, the IB-score, which measures the alignment between video and audio, demonstrates a notable improvement, surpassing a recent state-of-the-arts HunyuanVideo-Foley~\cite{shan2025hunyuanvideofoley} by 3.9 on the UnAV100 dataset. This reflects our model’s enhanced ability to capture and synchronize multimodal information effectively. Additionally, our method achieves consistently superior desynchronization scores, further emphasizing its robustness in handling complex audio-visual alignment tasks. These results collectively underscore the effectiveness of our approach in addressing real-world challenges in multimodal synchronization.
Moreover, we observe that autoregressive methods (\eg, V-AURA) struggle with length generalization, as evidenced by their comparatively poor performance among recent state-of-the-art techniques. Figure~\ref{fig:audio_specto} illustrates that previous methods fail to generate sound accurately aligned with the input video frames.
 On the LongVale dataset, our proposed method consistently outperforms state-of-the-art approaches by a substantial margin (0.23 on DeSync scores) compared to the second best performing method as shown in Table~\ref{tab:result_unav100_longvale}. Since LongVale contains samples with significantly longer durations (up to 7 minutes), this highlights that previous methods struggle with audio-video alignment and temporal synchronization when handling very long videos.  On VGGSound, where training and testing use identical durations, our proposed method performs on par with MMAudio~\cite{cheng2025mmaudio}, a strong baseline, and surpasses it on several key metrics (ISC scores), as shown in Table~\ref{tab:vggsound}. Please see our supplementary materials for generated samples and additional experiments.

\subsection{Analysis and Ablation Study}

\begin{table}[t!]
 \centering
    \resizebox{0.497\textwidth}{!}{
    \Large\addtolength{\tabcolsep}{-2pt}
\begin{tabular}{lccccc}
\toprule
\rowcolor{Lightgray} \multicolumn{6}{c}{\textbf{UnAV100}}\\
\textbf{Core Network} & \textbf{FD$_\textrm{VGG}$ $\downarrow$} & \textbf{FD$_\textrm{PANNs}$ $\downarrow$}    & \textbf{ISC$_\textrm{PANNs}$ $\uparrow$}   & \textbf{IB-Score $\uparrow$} & \textbf{DeSync $\downarrow$} \\
\toprule
Transformers &3.36 & 9.00   & 6.42    & 28.41 & 0.638 \\
Causal Mamba-2  &\textbf{2.28} & 9.18  &5.85   &33.32 &0.497 \\
 \rowcolor{aliceblue} Non-Causal Mamba-2 &3.35  &\textbf{5.87} &\textbf{7.62}   &\textbf{36.82} &\textbf{0.439} \\
 \toprule
\rowcolor{Lightgray} \multicolumn{6}{c}{\textbf{LongVale}}\\
\textbf{Core Network}  & \textbf{FD$_\textrm{VGG}$  $\downarrow$} & \textbf{FD$_\textrm{PANNs}$ $\downarrow$}    & \textbf{ISC$_\textrm{PANNs}$ $\uparrow$}   & \textbf{IB-Score $\uparrow$} & \textbf{DeSync $\downarrow$} \\
\toprule
Transformers  & 4.63  &\textbf{9.72}   & 2.95    & 18.65 &0.700 \\
Causal Mamba-2   & 6.62  &18.45 &3.74    &17.42 &0.743 \\
\rowcolor{aliceblue} Non-Causal Mamba-2  &\textbf{3.52}  &10.10    &\textbf{3.68}  &\textbf{30.62} &\textbf{0.438} \\
\bottomrule
\end{tabular}
}
\vspace{-0.1cm}
\caption{Ablation of the core networks of MMHNet by comparing among transformers, Causal Mamba-2 and Mamba-2. Evaluation is performed on UnAV100 and LongVale datasets.}
\vspace{-0.1cm}
\label{tab:result_ablation_corenetworks}
\end{table}

\vspace{0.1cm}
\noindent \textbf{Transformers Vs. Causal Mamba-2 Vs. Non-Causal Mamba-2.} We also provide a comparison with different types of core networks in Table~\ref{tab:result_ablation_corenetworks}. Transformers are done without positional embeddings attached to the tokens. Causal Mamba-2~\cite{dao2024mamba2} runs through tokens sequentially. Then, Non-causal Mamba-2~\cite{shi2024vssd} is used for our case to process long sequences and multimodal tokens more efficiently compared to causal Mamba-2.  

\begin{table}[t!]
 \centering
    \resizebox{0.497\textwidth}{!}{
    \Large\addtolength{\tabcolsep}{-2pt}
\begin{tabular}{lcccccccc}
\toprule
\rowcolor{Lightgray} \multicolumn{7}{c}{\textbf{UnAV100}}\\
\textbf{Variant}  & \textbf{FD$_\textrm{PANNs}$ $\downarrow$} & \textbf{FD$_\textrm{PASST}$ $\downarrow$}  & \textbf{ISC$_\textrm{PANNs}$ $\uparrow$} & \textbf{ISC$_\textrm{PASST}$ $\uparrow$} & \textbf{IB-Score $\uparrow$} & \textbf{DeSync $\downarrow$} \\
\hline
Non-Hierarchical   & 6.31   & 264.43  & 6.58  & 6.72 & 35.00 &0.621 \\
 \rowcolor{aliceblue} Hierarchical   & \textbf{5.87}   &\textbf{217.00}   &\textbf{7.62}  &\textbf{8.21} &\textbf{36.82} &\textbf{0.439} \\
\hline
\rowcolor{Lightgray} \multicolumn{7}{c}{\textbf{LongVale}}\\
\textbf{Variant}  & \textbf{FD$_\textrm{PANNs}$ $\downarrow$} & \textbf{FD$_\textrm{PASST}$ $\downarrow$}  & \textbf{ISC$_\textrm{PANNs}$ $\uparrow$} & \textbf{ISC$_\textrm{PASST}$ $\uparrow$} & \textbf{IB-Score $\uparrow$} & \textbf{DeSync $\downarrow$} \\
\hline
Non-Hierarchical  & 11.76   & 442.12  & 2.59  & 2.29 & 26.34 &0.669 \\
 \rowcolor{aliceblue} Hierarchical  & \textbf{10.10}   &\textbf{323.39} &\textbf{3.68} &\textbf{3.50}  &\textbf{30.62} &\textbf{0.438} \\
\bottomrule
\end{tabular}
}
\vspace{-0.1cm}
\caption{Hierarchical vs. non-hierarchical method comparison.}
\vspace{-0.5cm}
\label{tab:hierarchical_vs_non}
\end{table}

\vspace{0.05cm}
\noindent \textbf{Hierarchical Vs. Non-Hierarchical methods.}
We ablate on having the structure of models with tokens in the compressed space with tokens in the original space via routing mechanisms. We observe that the model with compressed space yields a better alignment between modalities in long audio generation forms in Table~\ref{tab:hierarchical_vs_non}. 

\begin{table}[t]
 \centering
    \resizebox{0.48\textwidth}{!}{
    \Large\addtolength{\tabcolsep}{-.5pt}
\begin{tabular}{cccccc}
\toprule
\textbf{Threshold}  & \textbf{FD$_\textrm{VGG}$ $\downarrow$}    & \textbf{FD$_\textrm{PANNs}$ $\downarrow$} & \textbf{ISC$_\textrm{PANNs}$ $\uparrow$} & \textbf{IB-Score $\uparrow$} & \textbf{DeSync $\downarrow$} \\
\hline
 
 0.3  & 3.24 &{8.13}  &{7.45} &33.44 &0.460 \\
 0.4  & 3.52 &{6.94}  &{8.61} &35.55 &0.431 \\
\rowcolor{Lightgray} 0.5  & \textbf{1.80} &\textbf{5.29}  &{8.10} &\textbf{36.27} &\textbf{0.410} \\
0.6  & 3.64 &6.91  &\textbf{8.66} &35.58 &0.426 \\
  0.7  & 15.03 &{33.06}  &{2.81} &0.02 &1.210 \\
\bottomrule
\end{tabular}
}
\vspace{-0.25cm}
\caption{Comparison of various threshold values on UnAV100.  }
\label{tab:threshold_unav100}
\end{table}

\noindent \textbf{Token selection thresholds.} As reported in Tab.~\ref{tab:threshold_unav100}, we systematically evaluated multiple threshold values to analyze their impact on overall performance. Among the tested settings, a threshold of 0.5 consistently produced the strongest results across all evaluation metrics. 

\begin{figure}[t!]
    \centering
    \vspace{-0.35cm}
 \subfloat{
        \includegraphics[width=0.24\textwidth]{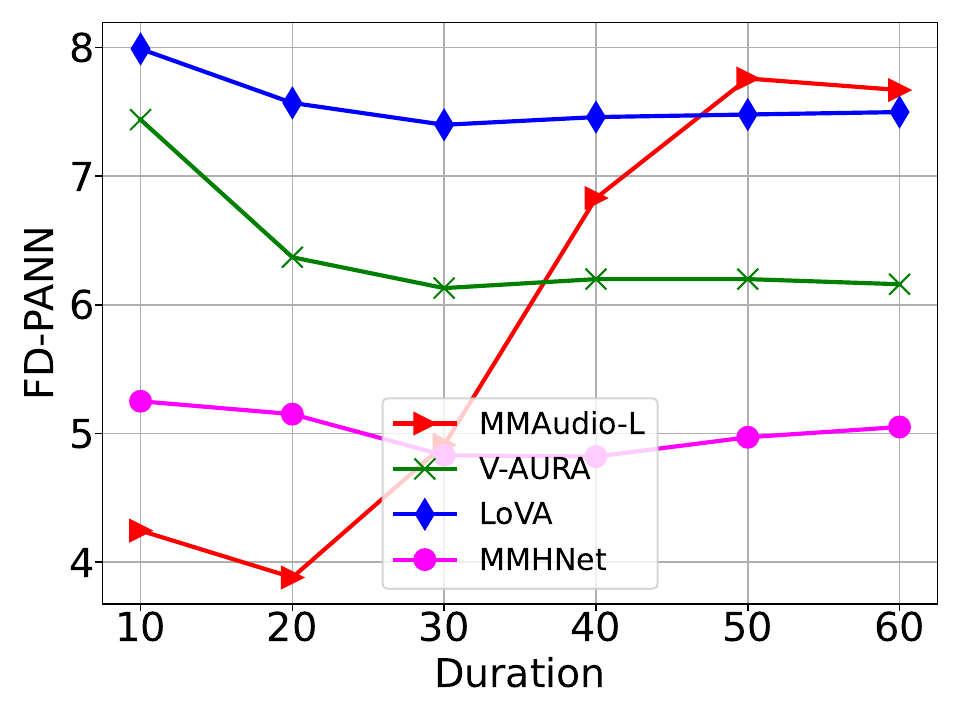}} 
 \subfloat{
        \includegraphics[width=0.24\textwidth]{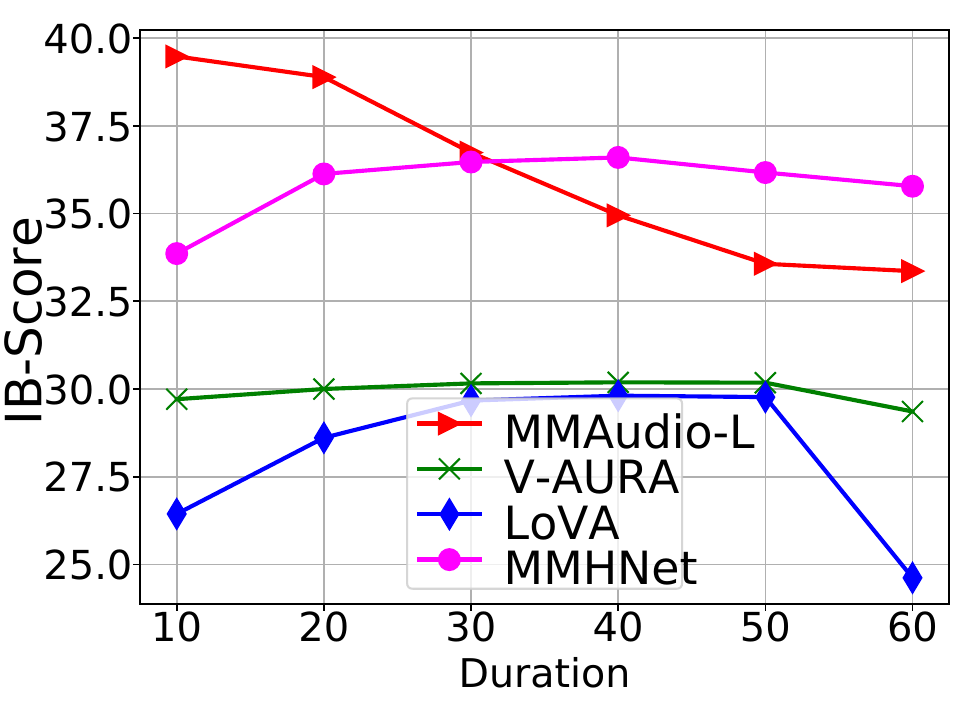}}  
\vspace{-0.38cm}
\caption{Comparison with past methods on various duration splits of audio-video data on UnAV100 (FD$_\textrm{PANNs}$ $\downarrow$ and IB-Score $\uparrow$).}
\vspace{-0.28cm}
     \label{fig:comparison_various_durations}    
\end{figure}

\vspace{0.1cm}
\noindent \textbf{Performance across various durations.}
We provide some analysis of different time durations to see the length generalization capability of our proposed method against the state-of-the-art method in V2A generation tasks (\eg, MMAudio~\cite{cheng2025mmaudio}). We show that past methods (\eg, MMAudio) fail to consistently maintain the performance across different durations. It is shown that FD$_\textrm{PANN}$ scores are plummeting to 3.5 points across video durations from 10 to 60 seconds, while our MMHNet can maintain the performance well. Also, MMHNet outperforms past methods (\eg, V-AURA and LoVA) in LV2A across durations as shown in Figure ~\ref{fig:comparison_various_durations}. 


\section{Conclusions}
This paper presents a hierarchical method so-called MMHNet, a novel framework for long-form video-to-audio generation that tackles the challenge of length generalization, training on short clips while generating high-quality, contextually aligned audio for much longer videos. MMHNet combines hierarchical modeling with a Non-Causal Mamba-2 architecture to overcome limitations of transformer-based models that rely on positional embeddings and struggle with long sequences. Hierarchical token routing and dynamic chunking efficiently align multimodal inputs (video, text, audio) while reducing complexity, and non-causal modeling ensures robust generalization.

\section{Acknowledgements}
We sincerely thank Kazuki Shimada and Ryosuke Sawata for constructive feedback to iteratively improve our work. 

{
    \small
    \bibliographystyle{ieeenat_fullname}
    \bibliography{main}
}



\end{document}